\documentclass[10pt,twocolumn,letterpaper]{article}

\usepackage[pagenumbers]{cvpr}      

\usepackage{amsmath}
\usepackage{algorithm}
\usepackage{algpseudocode}
\algrenewcommand\algorithmiccomment[1]{\hfill\small\textcolor{gray}{\textit{// #1}}}\algrenewcommand\algorithmiccomment[1]{\hfill\small\textcolor{gray}{\textit{// #1}}}

\usepackage[table]{xcolor}
\usepackage[dvipsnames]{xcolor}

\definecolor{CQColor}{rgb}{0.0,0.0,1.0} 

\definecolor{TSColor}{rgb}{0.5,0.0,0.8} 

\definecolor{CQRColor}{rgb}{1.0,0.0,0.0} 

\definecolor{mycolor_blue}{HTML}{E7EFFA}
\definecolor{mycolor_green}{HTML}{E6F8E0}
\definecolor{mycolor_gray}{HTML}{ECECEC}
\definecolor{pearDark}{HTML}{2980B9}
\definecolor{citecolor}{HTML}{2980b9}
\definecolor{linkcolor}{HTML}{c0392b}

\usepackage{threeparttable}

\definecolor{cvprblue}{rgb}{0.21,0.49,0.74}
\usepackage[pagebackref,breaklinks,colorlinks,allcolors=cvprblue]{hyperref}

\def\paperID{} 
\def\confName{CVPR}
\def\confYear{2026}

\title{MaskFocus: Focusing Policy Optimization on Critical Steps for \\ Masked Image Generation}
\author{Guohui Zhang$^{1}$ \quad Hu Yu$^{1}$ \quad Xiaoxiao Ma$^{1}$ \quad Yaning Pan$^{2}$ \quad Hang Xu$^{1}$ \quad Feng Zhao$^{1}$\thanks{Corresponding author}\\
$^{1}$University of Science and Technology of China $^{2}$Fudan University\\
}

\begin{document}

\let\oldtwocolumn\twocolumn
\renewcommand\twocolumn[1][]{
    \oldtwocolumn[{#1}{
    \begin{center}
    \includegraphics[width=\textwidth]{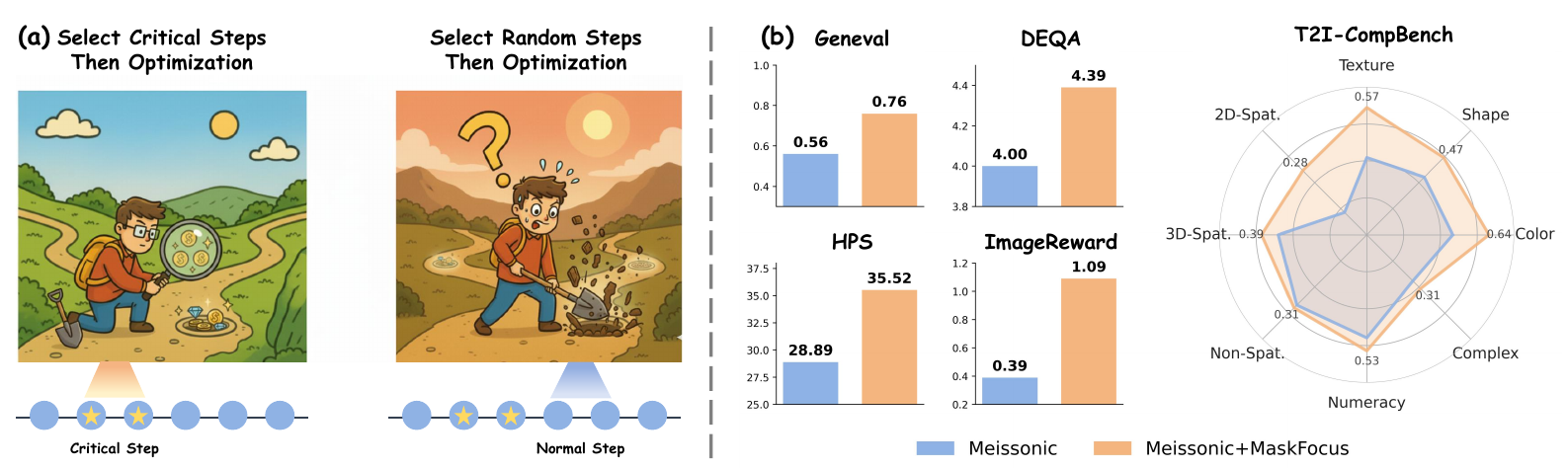}
    \captionof{figure}{(a) For masked generative models, certain steps in the sampling process are more valuable. The core of our method is to find these steps and perform policy optimization on them. (b) Our method achieves significant performance gains across multiple T2I benchmarks.}
    \label{fig:mainapp}
    \end{center}
    }]
}
\maketitle
\begin{abstract}
Reinforcement learning (RL) has demonstrated significant potential for post-training language models and autoregressive visual generative models, but adapting RL to masked generative models remains challenging.
The core factor is that policy optimization requires accounting for the probability likelihood of each step due to its multi-step and iterative refinement process. This reliance on entire sampling trajectories introduces high computational cost, whereas natively optimizing random steps often yields suboptimal results.
In this paper, we present \textbf{MaskFocus}, a novel RL framework that achieves effective policy optimization for masked generative models by focusing on critical steps.
Specifically, we determine the step-level information gain by measuring the similarity between the intermediate images at each sampling step and the final generated image.
Crucially, we leverage this to identify the most critical and valuable steps and execute focused policy optimization on them.
Furthermore, we design a dynamic routing sampling mechanism based on entropy to encourage the model to explore more valuable masking strategies for samples with low entropy.
Extensive experiments on multiple Text-to-Image benchmarks validate the effectiveness of our method.
Code is available at \url{https://github.com/zghhui/MaskFocus}
\end{abstract}    
\section{Introduction}
\label{sec:intro}
Recently, the field of visual generation has made significant progress, primarily driven by generative models like diffusion models (DMs)~\cite{ho2020denoising,song2020denoising,song2020score,lipman2022flow}, autoregressive generative models (AR)~\cite{sun2024autoregressive,liu2024lumina,wu2025janus}, and masked generative models (MGMs)~\cite{chang2022maskgit,he2022masked,chang2023muse,bai2024meissonic}. Among these paradigms, MGMs have emerged as an efficient and promising alternative, where all masked tokens are predicted in parallel at each step and only a subset is sampled based on confidence. This iterative non-autoregressive sampling process offers a significant speed advantage while maintaining high image quality. The representative work, Meissonic~\cite{bai2024meissonic}, demonstrates amazing capability in yielding high-fidelity samples comparable to SDXL~\cite{podell2023sdxl}.

Reinforcement learning (RL) has been pivotal for enhancing the reasoning capabilities of large language models (LLMs) in the post-training stage~\cite{guo2025deepseek,wang2025beyond,yue2025does}. Concurrently, recent efforts in AR~\cite{jiang2025t2i,zhang2025reasongen} and DMs~\cite{liu2025flow,zheng2025diffusionnft} have explored the potential of RL to improve model performance in visual quality and instruction following. However, extending these methods to MGMs remains challenging. Unlike AR, which is inherently compatible with LLMs, the coarse-to-fine iterative generation in MGMs requires step-wise probability estimation along the sampling trajectory. Mask-GRPO~\cite{luo2025reinforcement} reformulates the unmasking process as a multi-step decision-making problem and performs policy optimization over the entire trajectory, but this inevitably introduces a computational burden. MaskGRPO~\cite{ma2025consolidating} selects steps with higher mask ratios to enhance exploration and accelerate reward gains. However, we find that such non-dynamic selection underexplores the varying contributions of steps to the final image, resulting in suboptimal results. 

\begin{figure*}[htbp]
    \centering
    \includegraphics[width=\linewidth]{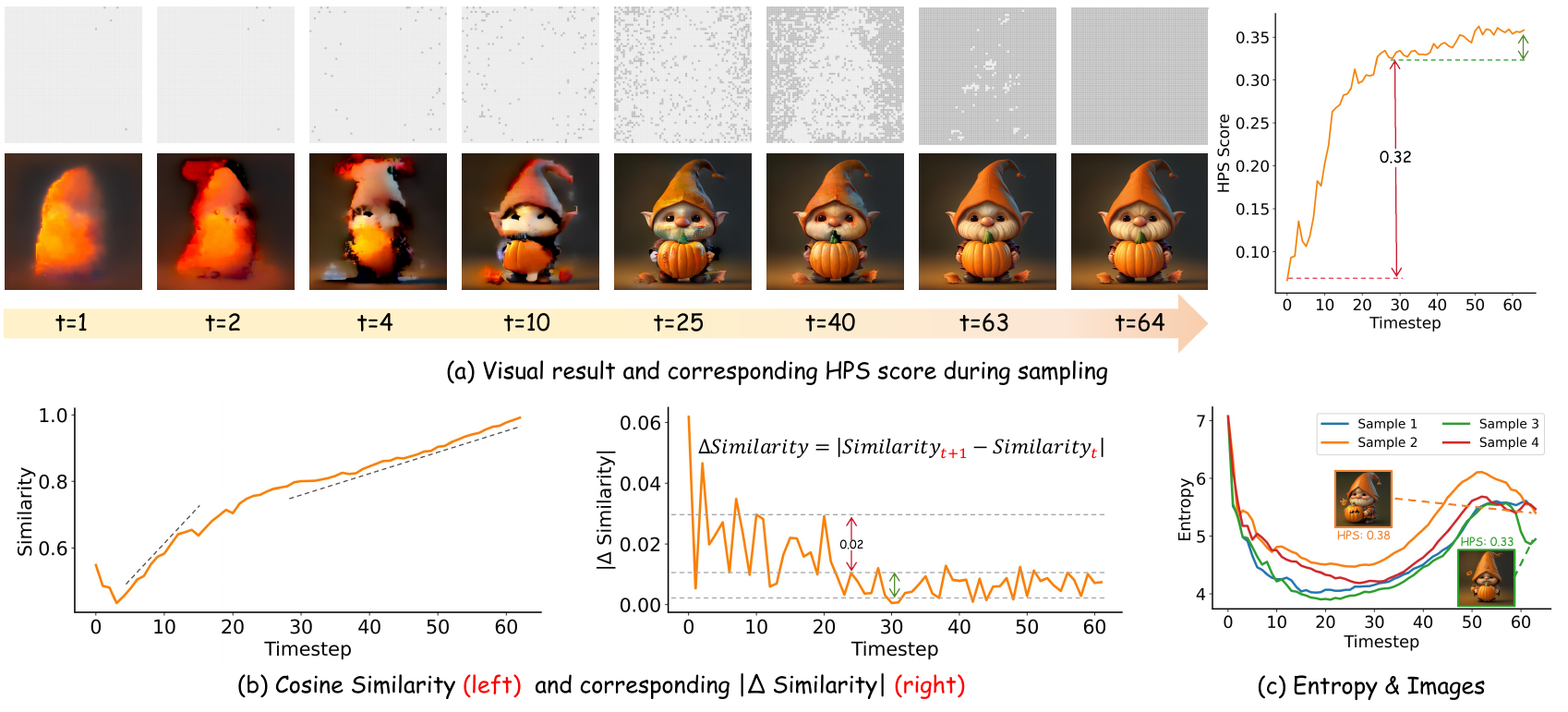}
    \caption{\textbf{Motivation of our method.} 
    (a) The masked tokens in the early steps determine the appearance and structure of the image, containing sufficient and effective information.
    (b) The \textbf{left} figure represents the cosine similarity $S_t$ between the image embedding $E_t$ at each step and the final embedding $E_T$. The \textbf{right} figure presents the absolute difference between consecutive steps, computed based on the similarity from the left figure. The image variation throughout the sampling process is not uniform. In particular, certain steps in the early stage have a more significant impact on the generated image.
    (c) Different samples exhibit different entropy trajectories during generation. Lower entropy implies more deterministic sampling, which limits exploration and makes it less likely to produce higher image quality.
    }
    \label{fig:motivation}
\end{figure*}

In this paper, we aim to identify the critical steps that are more valuable for policy optimization within the sampling trajectory. We observe that each masked token contains a valid estimation of the final generated image, as illustrated in Fig.~\ref{fig:motivation} (a). Moreover, early steps produce greater image changes, rapidly establishing the overall appearance and structure of the image, whereas the later steps are dedicated to refining local details. To further quantify this, we measure two metrics as shown in Fig.~\ref{fig:motivation} (b): \textbf{left:} the cosine similarity between the image embedding at each step and that of the final image; \textbf{right:} the absolute difference of this similarity between consecutive steps. Intuitively, the sampling trajectory is non-uniform, both between its different stages and between individual steps. We argue that this non-uniformity reflects the distinct contributions of each step to the final generated image and thus can serve as a potential measure for evaluating the value of sampling steps.

Based on these observations, we propose \textbf{MaskFocus} (see Fig.~\ref{fig:framework}), a novel RL framework that achieves effective policy optimization for MGMs by focusing on critical steps. For critical steps, we select the most valuable steps based on embedding similarity. For probability estimation of each critical step, we incorporate all masked tokens into the optimization objective. Additionally, we propose an entropy-based dynamic routing sampling mechanism that encourages the model to explore more valuable unmasking strategies. Our method further enhances the generative capability of the base model and outperforms existing masked-based RL methods. Extensive experiments on multiple text-to-image benchmarks validate the effectiveness of our approach. 
Our contributions are as follows:
\begin{itemize}
    \item Motivated by the observation that different sampling steps contribute unequally to the final image, we propose a critical-step selection strategy for MGMs to balance computational burden and performance in RL training.

    \item  We design an entropy-based dynamic routing sampling mechanism, achieving effective RL exploration for samples with low entropy. 

    \item Extensive experiments on multiple text-to-image benchmarks demonstrate the effectiveness of our method, pushing the generative capabilities of the mask-based generative model Meissonic to a new level. 
\end{itemize}

\section{Related Work}
\label{sec:related_work}

\subsection{Visual Generation}
Visual generation has witnessed remarkable advancements~\cite{yu2021vector,ho2020denoising,ma2024star,tian2024visual,yuan2025lumos}, primarily driven by three dominant paradigms: DMs, AR, and MGMs. DMs operate by iteratively transforming Gaussian noise into a coherent image through a reverse diffusion process.~\cite{rombach2022high} perform this process in latent space, paving the way for high-quality image generation.~\cite{peebles2023scalable,esser2024scaling} have further integrated the scaling capabilities of Transformers into diffusion models. In contrast to the iterative refinement, AR generate images sequentially by modeling the joint distribution of image tokens. LlamaGen~\cite{sun2024autoregressive} proposes an image generation model based on the Llama~\cite{touvron2023llama} architecture, while~\cite{liu2024lumina,zhou2024transfusion,xie2024show,chen2025janus,ai2025ming,chen2025blip3,wang2024emu3} unify visual and language tasks within a single multimodal autoregressive framework. Bridging the gap between these two approaches, MGMs~\cite{chang2022maskgit,chang2023muse,bai2024meissonic,li2024autoregressive,yu2025videomar} generate an image in an iterative manner, where all tokens for the image are produced simultaneously in each step, followed by iterative refinement conditioned on the previously generated tokens. 

\subsection{Masked Generative Generation}
Inspired by the bidirectional modeling of BERT~\cite{devlin2019bert} and the spatial nature of images, pioneering works~\cite{chang2022maskgit,zhang2021ufc} successfully apply bidirectional masked modeling to image generation, thereby validating the effectiveness of this paradigm. MaskGIT~\cite{chang2022maskgit} conducts an in-depth analysis of mask scheduling and introduces a highly effective iterative decoding strategy, which became foundational for subsequent work. Building upon this, Muse~\cite{chang2023muse} extends the masked image modeling to text-to-image synthesis, demonstrating exceptional visual quality and fine-grained prompt understanding. Meissonic~\cite{bai2024meissonic} further pushes the boundaries by employing a hybrid of unimodal and multimodal transformers and feature compression, successfully achieving high-quality and high-resolution image generation.

\subsection{Reinforcement Learning for Visual Generation.}
The success of Reinforcement Learning with Verifiable Rewards (RLVR) in LLMs~\cite{lambert2024tulu,yu2025dapo,zhang2025rlfr}, particularly Group Relative Policy Optimization (GRPO)~\cite{shao2024deepseekmath}, has led to a growing interest in leveraging these techniques to improve AR models~\cite{ma2025stage,jiang2025t2i,zhang2025group,zhang2025reasongen}.~\cite{liu2025flow,xue2025dancegrpo} successfully integrate online RL into flow matching models. Concurrently,~\cite{luo2025reinforcement} adapts GRPO to the masked generative paradigm by defining the iterative unmasking process as a transition probability. By introducing a novel sampling strategy,~\cite{ma2025consolidating} has significantly improved the image generation quality of multimodal discrete diffusion models. These pioneering efforts validate the efficacy of RLVR in optimizing the multi-step refinement processes inherent in modern image synthesis.

\section{Preliminary}
\begin{figure*}[htbp]
    \centering
    \includegraphics[width=\linewidth]{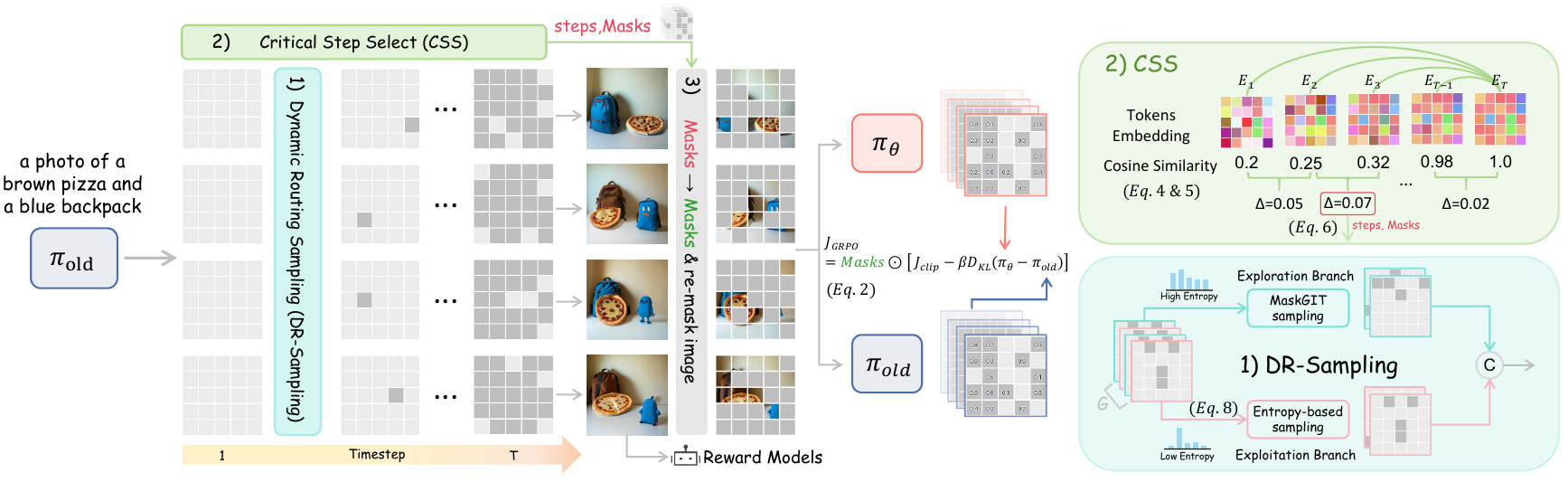}
    \caption{\textbf{Overview of our method.} 1) \textbf{Dynamic Routing Sampling (DR-Sampling)}. During the sampling process, we perform a more exploratory sampling strategy on low-entropy samples, while using normal sampling on high-entropy samples. 2) \textbf{Critical Step Select (CSS)}. Then, we determine the critical steps in the sampling trajectories and obtain the corresponding masks based on the cosine similarity between the intermediate embeddings and the final generated embedding. 3) We randomly shuffle masks and re-mask the generated tokens and predict the probabilities of these masked tokens to optimize the training objective (see left). Detail above procedures are in \textbf{Alg. \ref{alg:method}}.}
    \label{fig:framework}
\end{figure*}

\textbf{Masked Generative Models (MGMs)} contains three core components: an image tokenizer, a non-autoregressive transformer, and a predefined masking schedule. First, an image $I \in \mathbb{R}^{H \times W \times 3}$ is quantized into a sequence of discrete visual tokens $\{z_i\}^{N}_{1}$ by VQ-VAE. Then, a non-autoregressive transformer is trained to learn the joint distribution of these discrete tokens. During the training phase, a subset of tokens is randomly replaced with a special $\text{[MASK]}$ token. The model's objective is to predict all masked tokens $z_M$ in parallel, conditioned on the remaining visible tokens $z_V$ and text prompts:
\begin{equation}
\label{eq:logp}
    p(z_M | z_V) = \prod_{i \in M} p(z_i | z_V)
\end{equation}
The inference process begins with a canvas where all tokens are masked with the $\text{[MASK]}$ token and proceeds iteratively over $T$ steps. At each step $t$, the model predicts all masked tokens in parallel and keeps a subset of these tokens with the highest confidence scores. The remaining masked tokens are masked out and will be generated in subsequent steps.

The mask scheduling determines the dynamics of this iterative process. MaskGIT~\cite{chang2022maskgit} designs a mask scheduling function $\gamma(\cdot)$ to compute the mask ratio for the given latent tokens at each step $t$. To further enhance the quality and diversity of the generated images, recent works~\cite{ma2025towards,ma2025consolidating} have also explored more sophisticated sampling techniques, such as entropy-based or probability-based sampling strategies.

\textbf{Group Relative Policy Optimization (GRPO)~\cite{shao2024deepseekmath}} replaces the value function with a group relative reward to achieve simpler and more stable advantage estimation. For each input prompt $p$, the policy model $\pi$ generates $G$ outputs simultaneously, denoted as $\{o^i\}_{i=1}^{G}$. The reward model scores each sample to produce $R^i$, then the advantages $A^i$ are computed within each group. GRPO adopts a clipped objective similar to PPO~\cite{schulman2017proximal}. Besides, the optimization objective includes the KL penalty term between the reference model $\pi_{ref}$ and the policy model $\pi_{\theta}$. The objective of GRPO can then be written as:
\begin{equation}
\label{eq:grpo}
\begin{aligned}
\mathcal{J}  = 
\mathbb{E}_{\{o^i\} \sim \pi_{old}}
\Bigg[ \frac{1}{\sum_{i=1}^{G}|o^i|} \sum_{i=1}^{G}\sum_{t=1}^{|o^i|}
\color{black} \Bigg( \min\Big(r^i_{t}(\theta)\hat{A}^i,\\ 
    \text{clip}\left(r^i_{t}(\theta), 1 - \varepsilon, 1 + \varepsilon\right) \hat{A}^i \Big)
   - \beta {D}_{\text{KL}}(\pi_\theta \| \pi_{\text{ref}})
\Bigg)
\Bigg],
\end{aligned}
\end{equation}

\begin{equation}
\label{eq:advantage}
A^i = \frac{R^i - mean(\{R^i\}_{i=1}^{G})}{std(\{R^i\}_{i=1}^{G})}, r^i_{t}(\theta) = \frac{\pi_\theta(o^i_{t} \mid q, o^i_{<t})}{\pi_{\theta_{\text{old}}}(o^i_{t} \mid q, o^i_{<t})}
\end{equation}
where $r^i$ is importance ratio between the probabilities of $\pi_\theta$ and $\pi_{old}$. $D_{KL}$ is denotes the KL penalty.

\section{Motivation}
\textbf{Direct and effective probability estimation}. During the sampling process, the image's global structure and appearance are rapidly established in the early stages (steps 1-25), whereas subsequent steps (steps 25-64) are dedicated to refining local details and content, as shown in Fig.~\ref{fig:motivation} (a). Despite the fact that the MGMs' sampling strategy only selects the subset of tokens with the highest confidence at each step, we observe that the probability distribution of unselected masked tokens still exhibits a strong correlation with the final generated image. Moreover, since the pre-training objective of MGMs is to predict all masked tokens under the supervision of the ground truth, these masked tokens already contain information about the final image. Consequently, we directly leverage the probabilities of all masked tokens for policy optimization, following Eq.~\ref{eq:logp}.

\textbf{Non-equal importance for different steps.}
As the sampling process progresses, all masked tokens at each step increasingly align with those of the final generated image tokens. To quantify this intuition, we measure the cosine similarity $S$ between the image embedding at each step and that of the final image, as shown in Fig.~\ref{fig:motivation} (b). However, we find that the increase in similarity is not uniform: in the earlier steps, the similarity grows more rapidly, despite each step sampling fewer tokens than those in the later stages. To verify this, we calculate the absolute difference $\Delta S$ of the similarity between successive steps, which confirms this observation. Therefore, we argue that the contribution of each sampling step to the final image generation is not equal, especially for some earlier time steps, which contribute more to the generated image. We define this inter-step similarity change as the \textbf{information gain} of a given step $t$ to assess its value $v$. The detailed calculation is provided in Sec.~\ref{sec:CSS}. Consequently, this non-uniformity in contribution allows us to identify the critical steps within the sampling process.

\textbf{Balance between exploitation and exploration.} MGMs typically employ confidence-based sampling strategies, keeping masked tokens with the highest confidence as a priority. While this conservative strategy ensures a certain level of image generation quality. However, it tends to generate simple regions (e.g., background) in a large number of early steps, limiting the diversity and quality of the main visual subjects~\cite{ma2025towards}. We argue that this issue may stem from the low entropy of some sampling trajectories, as shown in Fig.~\ref{fig:motivation} (c). Low-entropy sampling trajectories indicate that the model adopts a deterministic sampling, while high-entropy trajectories do the opposite. Increasing the sampling temperature can introduce more randomness, encouraging the model to explore more unmasking locations. However, samples with high entropy are more sensitive to temperature compared to samples with low entropy, and overuse can easily lead to sampling instability or reward hacking. Therefore, we design an entropy-based dynamic routing sampling strategy to balance exploitation and exploration. We will discuss this in the Sec.~\ref{sec:sampling}.

\begin{algorithm*}[t]
\caption{MaskFocus}
\label{alg:method}

\textbf{Require:} Reference model $\pi_{\text{ref}}$, prompt dataset $\{q\}$, group size $G$, number of critical steps $K$. \\
\textbf{Initialize:} training policy $\pi_{\theta}\leftarrow\pi_{\text{ref}}$, critic-step buffer $\mathcal{C}\leftarrow \emptyset$, data buffer $\mathcal{D}\leftarrow \emptyset$
\begin{algorithmic}[1]
\While{not converged}
    \For {\text{each sampled prompt $q$}}
        \State Sample $G$ completions $o_i \sim \pi_{\text{old}}(\cdot|q)$, image embeddings $\{E\}^{1:T}$, masks $\{M\}^{1:T}$ in Sec.~\ref{sec:sampling} \Comment{DR-Sampling}
        \State For each $o_i$, compute advantage $A_i$ in Eq.~\ref{eq:advantage}  and information gain $\{\Delta S\}^{1:K}$ in Eq.~\ref{eq:information gain} 
        \State Select $K$ critical steps $\mathcal{C}$ in Eq.~\ref{eq:select step} and $\mathcal{D} \leftarrow\{o_k,k,M_k\},$ $k \in \mathcal{C}$ \Comment{Critical Step Selection}
    \EndFor
    \For {\text{each mini batch $\{o_k,k,M_k\} \in \mathcal{D}$}} 
            \State Randomly shuffle $M_k$ to get new mask $M^{'}_k$
            \State Construct masked completion $\hat{o}_i$ using $M^{'}_k$
            \State Compute log-likelihood $\pi_{\theta}$, $\pi_{old}$, $\pi_{ref}$ in Eq.~\ref{eq:logp} to estimate importance ratio $r^i$ and KL ${D}^{i}_{\text{KL}}$ in Eq.~\ref{eq:advantage}
            \State Compute policy optimization objective in Eq.~\ref{eq:grpo} and update $\pi_{\theta}$
    \EndFor
    \State Clear buffer $\mathcal{C}\leftarrow \emptyset$, $\mathcal{D}\leftarrow \emptyset$
\EndWhile
\end{algorithmic}
\textbf{Return:} $\pi_{\theta}$
\end{algorithm*}

\section{Method}
\label{sec:method}
Based on the above analysis, we propose \textbf{MaskFocus} (see Fig.~\ref{fig:framework} and Alg.~\ref{alg:method}), a novel RL framework that focuses policy optimization on critical steps. MaskFocus comprises two core designs: \textbf{Critical Step Select (CSS)} and \textbf{Dynamic Routing Sampling (DR-Sampling)}. The CSS is used to identify and select critical steps with the highest information gain during sampling. The DR-Sampling aims to balance exploration and exploitation within the group of samples, thereby enhancing model's policy learning capability.

\subsection{Critical Step Selection (CSS)}
\label{sec:CSS}
Firstly, we calculate the cosine similarity $S_t$ between the intermediate image embedding $E_t$ at each step $t$ and the embedding of the generated image $E_T$. Then, we calculate the absolute difference of similarity $\Delta S_t$ between consecutive steps as information gain $V_t$ at step $t$, which can be represented as:
\begin{equation}
\label{eq:information gain}
    S_t = CosSim(E_t, E_T),
\end{equation}
\begin{equation}
\label{eq:information gain}
    V_t = \left\vert\Delta S_t \right\vert = \left\vert S_{t+1} - S_{t}\right\vert
\end{equation}
We then select the top-$K$ steps with the highest information gain on all sampling steps as the critical steps:
\begin{equation}
\label{eq:select step}
    k = \arg\max_{t \in {1, ..., T-1}} \Delta V_t
\end{equation}
In this way, we focus the model's policy optimization on the steps that have a decisive impact on the final image and are pivotal in determining the trajectory of image quality. Furthermore, it also allows for more efficient allocation of the timestep budget.

\subsection{Dynamic Routing Sampling (DR-Sampling)}
\label{sec:sampling}
To encourage the model to learn more valuable sampling trajectories and achieve a better balance between exploration and exploitation, we design an entropy-based intra-group dynamic routing sampling strategy. Specifically, we first calculate the entropy $H_i$ of each sample before each sampling iteration, and then sort them according to their entropy values within the group.
\begin{equation}
H_{i} = - \sum_{v \in \mathcal{V}} p(v) \log p(v),
\end{equation}
where $V$ is the codebook, and $p(v)$ is the predicted probability of token.

\textbf{Exploitation branch.} For half of the samples with higher entropy, we route them to this branch to sample tokens. These high-entropy samples inherently possess greater uncertainty. Therefore, a standard confidence-based sampling strategy is applied to them to avoid introducing further uncertainty that could compromise training stability. 

\textbf{Exploration branch.} For half of the samples with lower entropy, we route them to this branch. Here, we employ an entropy-based dynamic temperature modulation strategy, following~\cite{ma2025towards}. This approach is designed to moderately perturb samples for which the model is already highly confident, encouraging the exploration of new possibilities.
\begin{equation}
    T_i = Te^{-\frac{H_{i,j}}{\alpha}} + \theta,
\end{equation}
where $H_{i,j}$ denotes the entropy at current token position $j$ of sample $i$, $T$ represents the maximum temperature, $\theta$ sets the lower bound, and $\alpha$ controls the decay rate of temperature with increasing entropy.

\begin{table*}[]
\centering
\caption{ \label{tab:generation_geneval} Quantitative comparison results on the GenEval benchmark. - represents unreported. The best result is in \colorbox{mycolor_green}{green}.}
\resizebox{\linewidth}{!}{
\begin{tabular}{l|ccccccc}
    \toprule
    \textbf{Method} & \textbf{Overall}$\uparrow$ & \textbf{Sing Obj.}$\uparrow$ & \textbf{Two Obj.}$\uparrow$ & \textbf{Counting}$\uparrow$ & \textbf{Color}$\uparrow$ & \textbf{Position}$\uparrow$ & \textbf{Color Attr.}$\uparrow$ \\
    \midrule
    \multicolumn{8}{c}{\textit{Diffusion-based Method}} \\
    \hline
    PixArt-$\alpha$~\citep{chen2024pixart}   & 0.32  & 0.98 & 0.50 & 0.44 & 0.80 & 0.08 & 0.07 \\ 
    SD1.5~\citep{rombach2022high} & 0.43 & 0.97 & 0.38 & 0.35 & 0.76 & 0.04 & 0.06 \\
    SD2.1~\citep{rombach2022high} & 0.50 & 0.98 & 0.51 & 0.44 & 0.85 & 0.07 & 0.17 \\
    LDM~\citep{rombach2022high} & 0.37 & 0.92 & 0.29 & 0.23 & 0.70 & 0.02 & 0.05 \\
    SDXL~\citep{podell2023sdxl} & 0.55 & 0.98 & 0.74 & 0.39 & 0.85 & 0.15 & 0.23 \\
    SD3.5-M~\citep{esser2024scaling} &  0.63 & 0.98 & 0.78 & 0.50 & 0.81 & 0.24 & 0.52 \\  
    DALL-E 3~\citep{betker2023improving} & 0.67 & 0.96 & 0.87 & 0.47 & 0.83 & 0.43 & 0.45 \\
    FLUX.1-dev~\citep{Flux} & 0.66 & 0.98 & 0.81 & 0.74 & 0.79 & 0.22 & 0.45 \\
    \hline
    \multicolumn{8}{c}{\textit{AR-based method}} \\
    \hline
    LlamaGen~\citep{sun2024autoregressive} & 0.32 & 0.71 & 0.34 & 0.21 & 0.58 & 0.07 & 0.04 \\
    JanusFlow~\citep{ma2025janusflow} & 0.63 & 0.97 & 0.59 & 0.45 & 0.83 & 0.53 & 0.42 \\    
    Janus-Pro-1B~\citep{chen2025janus} & 0.73 & 0.98 & 0.82 & 0.51 & \colorbox{mycolor_green}{0.89} & \colorbox{mycolor_green}{0.65} & 0.56 \\
    SimpleAR-1.5B~\citep{chen2025janus} & 0.63 & - & 0.90 & - & - & 0.28 & 0.45 \\   
    Chameleon~\citep{team2024chameleon} & 0.39 & - & - & - & - & - & - \\  
    Emu3 (+Rewriter)~\citep{sun2023emu} & 0.66 & 0.99 & 0.81 & 0.42 & 0.80 & 0.49 & 0.45 \\    
    \hline
    \multicolumn{8}{c}{\textit{Masked Generative Model}} \\
    \hline
    Show-o~\citep{xie2024show}& 0.68 & 0.98 & 0.80 & 0.66 & 0.84 & 0.31 & 0.50 \\
    Show-o+PARM~\citep{guo2025can} & 0.69 & 0.97 & 0.75 & 0.60 & 0.83 & 0.54 & 0.53 \\
    Mask-GRPO~\citep{luo2025reinforcement} & 0.73 & 0.99 & 0.90 & 0.69 & 0.85 & 0.35 & \colorbox{mycolor_green}{0.59} \\ 
    Meissonic~\citep{bai2024meissonic} & 0.54 & 0.99 & 0.66 & 0.42 & 0.86 & 0.10 & 0.22 \\
    \quad+ MaskGRPO~\citep{ma2025consolidating} & 0.73  &  \colorbox{mycolor_green}{1.00} &  0.87 & 0.83  & 0.87  & 0.39 &  0.48 \\ 
    \quad+ MaskFocus (our) & \colorbox{mycolor_green}{0.76} & \colorbox{mycolor_green}{1.00} & \colorbox{mycolor_green}{0.91} & \colorbox{mycolor_green}{0.85} & 0.87 & 0.42 & 0.54 \\    
    \bottomrule
\end{tabular}
}
\end{table*}

\section{Experiments}
\label{sec:experiment}
In this section, we empirically evaluate the effectiveness of our method to improve the performance of MGMs on two representative tasks. (1) Compositional Image Generation. This task is designed to evaluate the model's ability to control attributes, including counting, color, and position tasks. We report the results on the GenEval~\cite{ghosh2023geneval} and T2I-CompBench~\cite{huang2023t2i}. (2) Image Quality and Preference Alignment. This task evaluates the quality and aesthetics of images generated by the model. We report DEQA-Score~\cite{you2025teaching} and ImageReward~\cite{xu2023imagereward}, HPS~\cite{wu2023human}, and PickScore~\cite{kirstain2023pick} on DrawBench~\cite{saharia2022photorealistic}.

\subsection{Experimental Setup}
\textbf{Training Setup}. Our experiments are based on Meissonic~\cite{bai2024meissonic}, an open-source masked generative model with $1024\times1024$ resolution. The CFG is 5 in both training and inference. The sampling step is 64, and the number of critical steps is 6. More hyperparameters can be found in the supplementary materials.

\textbf{Training Data}. For compositional image generation, we utilize 50k Geneval-Style prompts sourced from Flow-GRPO~\cite{liu2025flow}. These prompts are characterized by being short and simple and are related to control attributes. These prompts do not overlap with the official GenEval prompts. For the preference alignment task, we use 10k training samples from the HPS v2 dataset~\cite{wu2023human}, which are known for their complexity and diversity.

\textbf{Reward}. For compositional image generation, we utilize both CLIP~\cite{radford2021learning} and GenEval reward as reward. For the preference alignment task, we use HPS as the reward.

\textbf{Baseline.} 
We primarily compare our method with MaskGRPO~\cite{ma2025consolidating} to demonstrate the effectiveness of our critical step selection strategy. For Mask-GRPO~\cite{luo2025reinforcement}, we only report results from their original paper, since the code is not yet open-source.

\begin{figure*}[htbp]
    \centering
    \includegraphics[width=\linewidth]{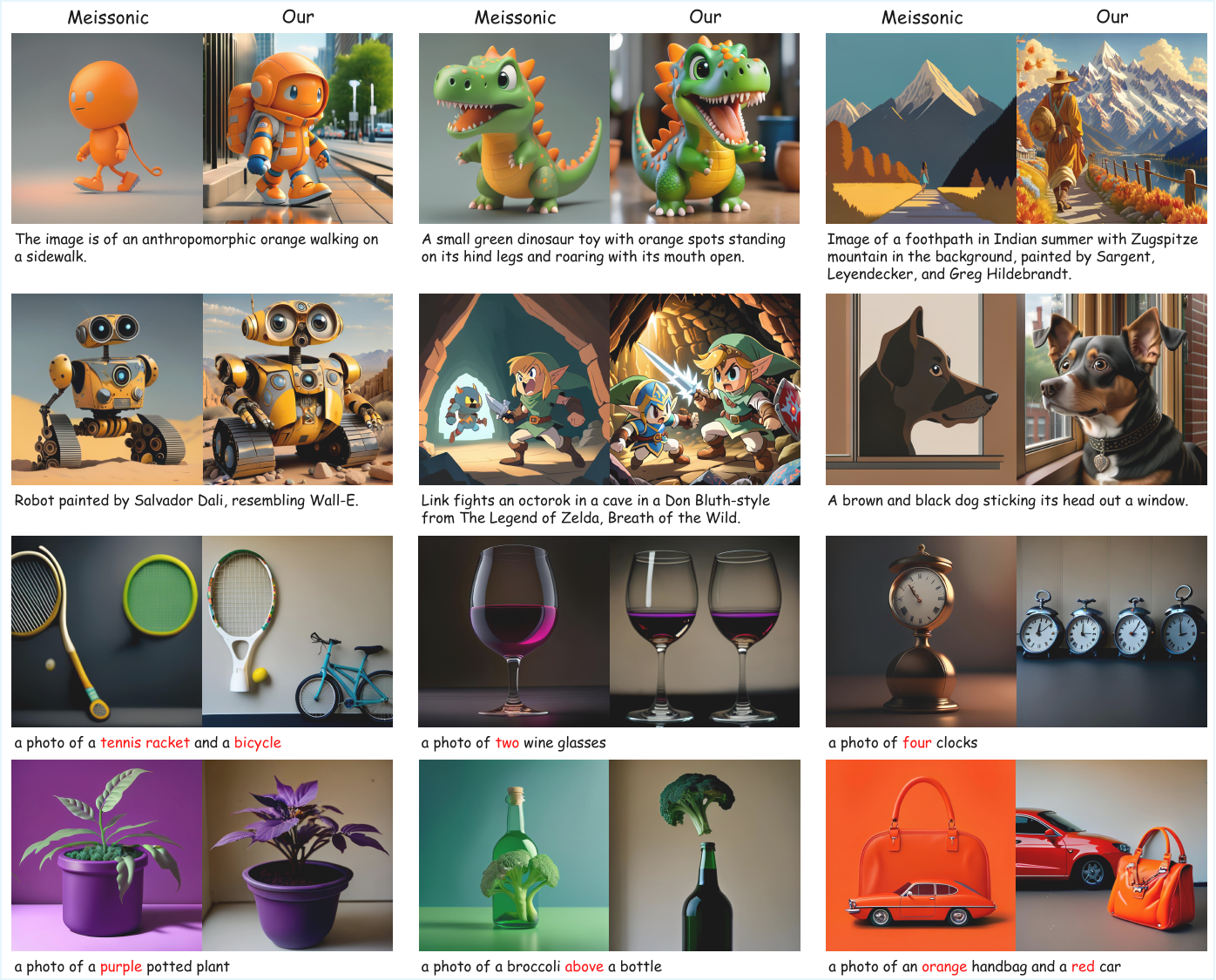}
    \caption{\textbf{Qualitative Comparison.}  Our approach demonstrates superior performance in image quality and human preference (top two rows), as well as in instruction-following tasks involving Counting, Colors, Attribute Binding, and Position (bottom two rows).}
    \label{fig:visual_result}
\end{figure*}

\begin{table}[htbp]
\centering
\caption{ \label{tab:quality} Quantitative comparison results on the human preference metrics. The best result is in \colorbox{mycolor_green}{green}.}
\resizebox{\linewidth}{!}{
\begin{tabular}{l|cccc}
    \toprule
    \textbf{Method} & \textbf{DEQA}$\uparrow$ & \textbf{PickScore}$\uparrow$ & \textbf{HPS}$\uparrow$ & \textbf{ImageReward}$\uparrow$ \\
    \midrule
    SDXL~\citep{podell2023sdxl}  & 4.17 & 22.27  & 30.06  & 0.56 \\ 
    SD3.5-M~\citep{esser2024scaling}   & 4.24 & 22.50  & 30.17 & 0.98\\ 
    FLUX.1-dev~\citep{Flux}  & 4.37  & \colorbox{mycolor_green}{22.97} & 31.13 & 1.06\\ 
    Meissonic~\citep{bai2024meissonic}   & 4.00  & 21.63 & 28.89 & 0.39\\ 
    \quad+ MaskGRPO~\citep{ma2025consolidating} & 4.35 & 22.34 & 35.48 & 1.06\\
    \quad+ MaskFocus (our) & \colorbox{mycolor_green}{4.39} & 22.39 & \colorbox{mycolor_green}{35.52} & \colorbox{mycolor_green}{1.09}\\
    \bottomrule
\end{tabular}
}
\end{table}

\subsection{Main Results}
\textbf{Quantitative comparison.} As shown in the Table.~\ref{tab:generation_geneval}, we compare our method with leading text-to-image diffusion and autoregressive models on the Geneval benchmark.Our approach achieves notable improvements on sub-tasks such as counting and position, and outperforms MaskGRPO. To further validate the generalization capability of our method, we conduct additional experiments on the T2I-Compbench, as illustrated in Fig.~\ref{fig:mainapp} (b). We exhibit excellent generalization performance across multiple subtasks. Notably, our method further improves the image quality of Meissonic on multiple metrics, achieving results comparable to the FLUX, as shown in Table.~\ref{tab:quality}.

\textbf{Qualitative comparison.} As illustrated in Fig.~\ref{fig:visual_result}, we provide samples generated by sourced Meissonic and our method across multiple prompts. The top two lines further illustrate that our method consistently produces visually superior results with higher fidelity and richer detail. The bottom two lines explain that our method provides more accurate instruction alignment for prompts.

\subsection{Ablation Results}
\textbf{Ablation Study.} As shown in Table~\ref{tab:ablation}, we conduct ablation studies to evaluate the contribution of each key component in our framework. \textbf{Removing the Critical step selection} and randomly selecting some early steps instead leads to a performance drop across all metrics, which indicates that our selected steps are more valuable for the RL training. 

\begin{table}[htbp]
\centering
\caption{ \label{tab:ablation} Ablation Result.}
\resizebox{\linewidth}{!}{
\begin{tabular}{l|ccc}
    \toprule
    \textbf{Method}& \textbf{Geneval}$\uparrow$  & \textbf{DEQA}$\uparrow$ & \textbf{PickScore}$\uparrow$ \\
    \midrule
    MaskFocus (Our)  & 0.76  & 4.39  & 22.39 \\ 
    \quad w/o Critical Step Selection  & 0.72  & 4.34 & 22.34\\ 
    \quad w/o Dynamic Routing Sampling   & 0.74  & 4.35 & 22.31\\ 
    \bottomrule
\end{tabular}
}
\end{table}

\textbf{Removing dynamic routing sampling.} We observe a more significant drop in image quality, while the GenEval metric is less affected. This demonstrates that our dynamic sampling achieves efficient exploration and further improves performance. We compare the performance of different training processes in more detail in the figure. Notably, our full method achieves the current state-of-the-art results on all metrics.

\begin{figure*}[htbp]
    \centering
    \includegraphics[width=\linewidth]{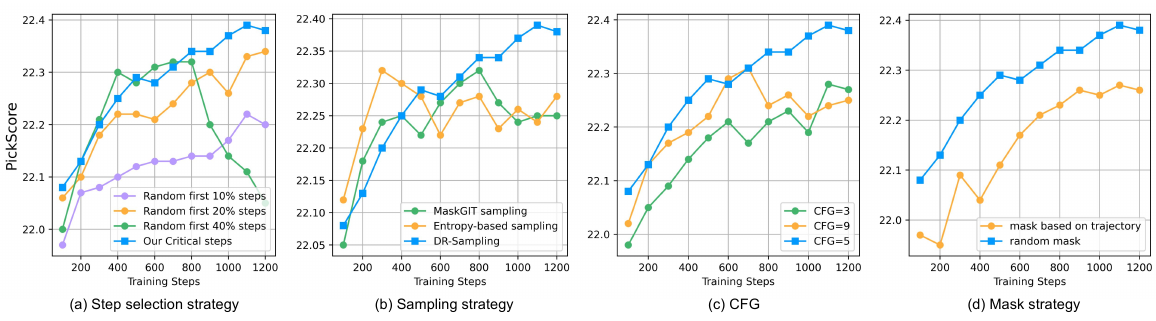}
    \caption{\textbf{More comparison results on step selection strategy, sampling strategy, CFG, and mask strategy.} }
    \label{fig:ablation}
\end{figure*}

\subsection{More Discussions}
\textbf{Different step selection strategies.} 
We further compare various selection strategies. As shown in Fig.~\ref{fig:ablation} (a), we observe that randomly selecting steps within the first 40\% steps results in performance degradation during the later training stages. Further analysis reveals that this configuration is prone to reward hacking. We attribute this phenomenon to the model overfitting to minor refinement details, which allows it to more easily exploit loopholes in the reward function. Moreover, while training on steps from a 20\% interval yields a significant advantage over using only the earliest steps, it still underperforms our proposed method. This implies that although early steps are effective, there are still better step selections available. Our approach of training on critical steps facilitates a more stable and effective RL training.

\textbf{Different sampling strategy.} We compare MaskGIT sampling, Entropy-based sampling for all samples, and our dynamic routing sampling, as shown in Fig.~\ref{fig:ablation} (b). With MaskGIT sampling, limited exploration hinders  performance improvements. In contrast, applying Entropy-based sampling to all samples leads to unstable training, especially in the later training stage. Our dynamic routing sampling effectively balances both approaches.

\textbf{The influence of CFG.}
Meissonic's default CFG is 9. However, we find that higher and lower CFG are suboptimal, as shown in Fig.~\ref{fig:ablation} (c). The former induces significant shifts in the model's sampling distribution and the latter fails to guarantee image quality during sampling. Conversely, a moderate CFG setting yields superior training performance. Consequently, we employ a moderate CFG during training and inference.

\textbf{Mask stagery.} We investigate the impact of different masking strategies on probability estimation, as illustrated in Fig.~\ref{fig:ablation} (d). A straightforward approach is to re-mask the image using the generated masks during sampling. However, our experiments show that this strategy is suboptimal compared to regenerating the masks. We argue that this trajectory-based masking method exhibits significant probability estimation errors under the off-policy training paradigm, leading to excessively large KL penalty between the policy model and the reference model. This could be a focus of future research.

\begin{figure}[htbp]
    \centering
    \includegraphics[width=\linewidth]{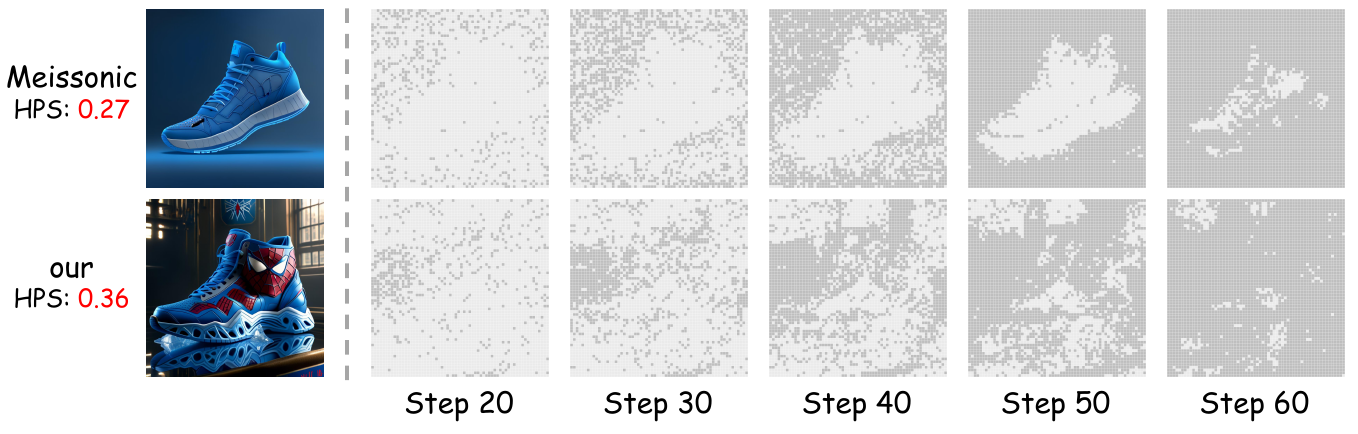}
    \caption{\textbf{Pre-/Post-RL sampling trajectories.} }
    \label{fig:image_mask}
\end{figure}

\textbf{Sampling differences between the original model and the RL-trained model.} We analyze the sampling process of the original model and the model after RL training, as shown in Fig.~\ref{fig:image_mask} (d). The original model's sampling process is singular and primarily proceeds from background to subject, where background content requires more sampling steps. This often leads to degraded quality in the main subjects~\cite{ma2025towards}. In contrast, models trained with RL exhibit a more diverse sampling process and higher-quality images.
\section{Conclusion}
\label{sec:conclusion}
In this paper, we propose MaskFocus, a novel RL framework for masked generative models that focuses policy optimization on critical sampling steps and incorporates entropy-based dynamic routing sampling to enhance exploration. Our approach effectively identifies the value of each step based on image embedding similarity and selects the most valuable steps as critical steps, thereby balancing computational efficiency and performance. Our dynamic routing sampling encourages the model to steadily explore more valuable mask strategies. Extensive experiments on multiple text-to-image benchmarks demonstrate that MaskFocus advances the generative capabilities of MGM models.

{
    \small
    \bibliographystyle{ieeenat_fullname}
    \bibliography{main}
}


\end{document}